\documentclass[letterpaper, 10 pt, conference]{ieeeconf}  

\IEEEoverridecommandlockouts                              

\usepackage{amsmath} 

\usepackage{hyperref}
\usepackage{xcolor}
\usepackage{graphicx}
\usepackage{algorithm} 
\usepackage[noend]{algpseudocode}
\algdef{SE}[SUBALG]{Indent}{EndIndent}{}{\algorithmicend\ }%
\usepackage{makecell}
\usepackage{amsfonts}

\usepackage{stfloats}

\usepackage[T1]{fontenc} 

\title{\LARGE \bf Pack it in: Packing into Partially Filled Containers Through Contact
}

\author{David Russell$^{1}$, Zisong Xu$^{2}$, Maximo A. Roa$^{3}$, Mehmet Dogar$^{1}$
\thanks{$^{1}$School of Computer Science, University of Leeds, UK.
{\tt\small \{scsdr, m.r.dogar\}@leeds.ac.uk}}
\thanks{$^{2}$School of Intelligence Science and Technology, Nanjing University, China. {\tt\small zisongxu@nju.edu.cn}}
\thanks{$^{3}$German Aerospace Centre (DLR), Germany.}
\thanks{This research has received funding from the UK Engineering 
and Physical Sciences Research Council under the grant EP/V052659/1.}
\thanks{For the purpose of open access, the authors have applied a
Creative Commons Attribution (CC BY) license to any Author Accepted
Manuscript version arising. Please see attached \href{https://www.youtube.com/watch?v=tu5DuR0WgOU}{video}.}
}

\begin{document}

\maketitle
\thispagestyle{empty}
\pagestyle{empty}

\begin{abstract}

The automation of warehouse operations is crucial for improving productivity and reducing human exposure to hazardous environments. One operation frequently performed in warehouses is bin-packing where items need to be placed into containers, either for delivery to a customer, or for temporary storage in the warehouse. Whilst prior bin-packing works have largely been focused on packing items into empty containers and have adopted collision-free strategies, it is often the case that containers will already be partially filled with items, often in suboptimal arrangements due to transportation about a warehouse. This paper presents a contact-aware packing approach that exploits purposeful interactions with previously placed objects to create free space and enable successful placement of new items. This is achieved by using a contact-based multi-object trajectory optimizer within a model predictive controller, integrated with a physics-aware perception system that estimates object poses even during inevitable occlusions, and a method that suggests physically-feasible locations to place the object inside the container.

\end{abstract}

\section{INTRODUCTION}

\begin{figure}[t]
    \centering
    \includegraphics[width=0.98\linewidth]{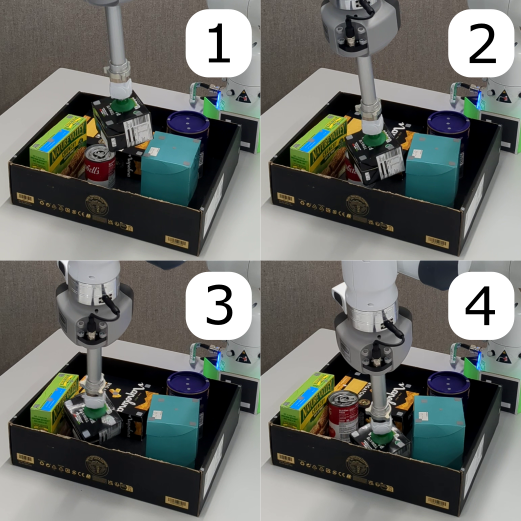}
    \caption{Snapshots of the proposed packing through contact system operating on real robotic hardware. The system uses contact-based interactions between the inserted object and objects in the container to clear space and place it within the container. Please also see attached video.}
    \label{fig:packing_system}
\end{figure}

This work focuses on the problem of packing an object into a container by physically interacting with the objects already inside. We present an example execution by our proposed system in Fig.~\ref{fig:packing_system}, where the robot is tasked with packing the gripped object inside an already partially filled container. There is, however, no continuous free space large enough to  place the object; therefore, the robot needs to contact and move the existing objects to create space. The robot contacts the red can, pushing it to the left, while simultaneously nudging the left-most green box. The robot then also nudges the rightmost blue box, while inserting the gripped object to its place.  

This is an important robotic manipulation problem, particularly in warehouse automation.
Amazon Robotics' recent work on \textit{stowing} \cite{hudson2025stow} demonstrates a system that inserts objects into partially filled containers/shelves for storage inside the warehouse, with the help of a specially designed conveyor-gripper mechanism integrated with a thin planar tool.  
Bin-packing \cite{wang2019stable,wang2021dense,agarwal2020jampacker,xiong2023towards} systems insert objects sequentially into a container, to prepare them for shipping to customers. Our proposed approach has the potential to extend the capability of such warehouse systems, by enabling them to explicitly plan for contact-based interactions between the inserted object and other objects in the container. This problem, however, comes with several challenges, three of which we address in this work. 

First, is the challenge of motion planning. Given a target placement pose for the object, if there is a collision-free path to move the object to that pose, then the problem can be solved using traditional motion planners. Many existing systems use such a collision-free approach, by carefully packing objects one-by-one and trying to maintain collision-free space for the next object to be inserted \cite{wang2019stable,wang2021dense,agarwal2020jampacker,xiong2023towards}. While this is a good strategy in general, in practice, objects in a container move and shift, making it extremely difficult to always guarantee that the next object can be inserted without contacting any other object, especially as the container becomes crowded. The next object to be inserted is also not always known beforehand (called `online bin-packing' \cite{ali2022online}), as also in the case of the Amazon stowing \cite{hudson2025stow} problem. This requires some interaction with the objects in the container to clear space. Song et al. \cite{song2019object} and Shome et al. \cite{shome2021tight} address this problem by planning specific non-prehensile actions inside the container to rearrange the existing objects, either, to create space for the new object or to achieve tight packing. The new object is then inserted in a collision-free manner. 
However, the proposed non-prehensile actions in these works are  specific to cuboid objects. Instead, we propose a general contact-based trajectory optimization formulation, based on iLQR \cite{li2004iterative,tassa2012synthesis}, that can interact with objects of any shape. Furthermore, Song et al. \cite{song2019object} and Shome et al. \cite{shome2021tight} address scenarios where the robot needs to plan two separate phases of interaction, one phase for rearrangement and a second phase for collision-free insertion. While this is sometimes necessary during packing, and separate phases are sometimes even employed by human packers, there are also situations where rearrangement and insertion can be performed simultaneously. In this work, we focus on developing such a skill for robots, which can improve the time efficiency when separate actions are not necessary.

The second challenge is the robust execution of the planned motion. Contact-based trajectory optimization, as explained above, requires a physics model of how the contacted objects move. These models, however, are never perfect, and the real objects' motion differ from the predictions of the physics model. This requires adapting the robot's motion during the interaction. We address this by using a model predictive control (MPC) approach,  which continuously tracks all the objects' poses, and re-optimizes the trajectory accordingly. However, tracking the objects' poses (i.e., perception) is inherently difficult in these settings, as cluttered environments lead to significant occlusions, especially when the robot, or the gripped object, is between any vision system and the container. In our system, we use marker-based tracking  (OptiTrack) to track objects, but even such a marker-based method regularly loses objects, due to the high-degree of occlusions. Therefore, we integrate our MPC with a simplified physics-aware perception system based on work from Xu et al. \cite{xu2025tracking} that uses physics predictions to estimate the poses of objects within the container even when they are not visible to the tracking system.

The third, and final, challenge we address is to find a feasible \textit{placement pose} for the new object when no collision-free location exists within the container. When discussing the first two challenges above, we assumed that a target placement pose was given. However, choosing such a pose by itself is an important problem: while a good choice can make the motion planning problem much easier, a bad choice can cause the whole system to fail. The key insight we have is that, finding a good placement pose also requires physics-based reasoning.
Therefore, we propose a placement planner that observes the current state of the partially filled container and suggests feasible packing poses for a new object. Candidate placements are evaluated by simulating physical interactions within the container, and the planner selects the feasible configuration that minimally disturbs the existing arrangement.

In summary, we:

\begin{itemize}
    \item formulate the packing problem as a physics-based trajectory optimization in clutter problem to be solved with iLQR. This formulation is able to use efficient contact-based interactions to clear space inside a container and pack a new object in one continuous action;
    \item propose a physics-informed placement planner that  outputs a physically-feasible pose to pack the new object;
    \item integrate the above in an MPC framework with a physics-aware perception system to assist in scenarios when objects are obscured from view.
\end{itemize}

We perform extensive real-robot experiments to measure the performance of our system, along with an analysis of the failure modes. Please see attached video for example runs. 

\subsection{Other related work}\label{sec:related_work}

We already discussed some related works above. Additionally, Moroni et al. \cite{moroni2024robotic} consider bin packing where objects need to be inserted into a bin tightly. Using custom hardware, they propose a particular insertion strategy that considers the geometry of the inserted object and objects within the container. Their strategy then follows a geometric path where it tilts the inserted object into a gap and pulls back a surrounding object to make a gap large enough for object insertion. The key difference between our insertion strategy and their is ours can implicitly consider objects of different geometries (instead of purely cuboid objects) as well as consider complicated multi-object interactions implicitly by leveraging MPC and a general purpose physics simulator.

More broadly, our work formulates the bin packing problem as a trajectory optimization through clutter task. Trajectory optimization through clutter is often difficult due to the long optimization times that occur in high-dimensional problems, such as manipulation in clutter. A variety of works have investigated using trajectory optimization with a general purpose physics simulator for manipulation in clutter tasks, such as reaching through clutter \cite{kitaev2015physics,papallas2020online,agboh2018real}. Importantly, these works use computationally expensive long-horizon planning and therefore do not consider Model Predictive Control (MPC) on real robotic hardware. 


There is also a rich body of works that consider the bin packing problem (whether it is online or offline) without considering the robotic manipulator motion \cite{martello2000three,dosa2013first,wang2010two}. In this work, our focus is on performing packing with robots.

\section{PROBLEM FORMULATION} \label{sec:problem_formulation}

We consider the problem of object packing into a cluttered container. The robotic manipulator is described by its joint configuration vector $
\mathbf{q} = [q_1, q_2, \ldots, q_n]^\top$, with $n$ being the number of joints. The container is modelled as a bounded, static volume in 3D space, denoted as $\mathcal{C} = [0, W] \times [0, L] \times [0, H] \subset \mathbb{R}^3$, where \(W, L, H > 0\) represent its width, length, and height. There is a height threshold $h_{max}$ which objects need to be below to be considered inside the container.

We determine that an object $O$ at pose $T \in SE(3)$ is contained in the container if the volume of $O$, lies entirely within $C$ and its height is below the height threshold $t_z(T) < h_{max}$, where $t(T) \in \mathbb{R}^3$ denotes the translation component of a pose $T$, and $t_z(T)$ its vertical coordinate. The existing contents of the container are given by a set of \(N\) rigid objects, $\mathcal{O} = \{ O_1, O_2, \ldots, O_N \}$, where each object $O_i$ has known geometry and a 6D pose $T_i \in SE(3)$, specifying its position and orientation in the world frame, with $R_i \in SO(3)$ denoting the rotational component.

Given a new object $O_{\text{new}}$, the goal is to contain the object within the volume of $\mathcal{C}$. When sufficient continuous free space exists, this reduces to a standard collision-free motion planning problem. However, we focus on the more challenging case where there is insufficient free space, requiring rearrangement of existing objects $\mathcal{O}$ within the container to accommodate $O_{\text{new}}$.

\section{METHOD} \label{sec:method}

Sec. \ref{sec:traj_opt_formulation} formulates the problem of packing through clutter as one to be solved via trajectory optimization. It is formulated to move $O_{\text{new}}$ to some goal location, being able to use non-prehensile manipulation to displace objects within the container as required. Sec. \ref{sec:MPC_system_design} integrates our physics-based trajectory optimization methods into an MPC framework to account for physical inaccuracies in the simulator to enable robust execution on real robotic hardware. Finally, to determine a suitable goal pose for $O_{\text{new}}$, we propose a placement planner in Sec. \ref{sec:placement_planner} which observes the state of a partially filled container and proposes a suitable packing location that is dynamically feasible considering the constraints of objects within the container.

\subsection{Trajectory Optimization Formulation} \label{sec:traj_opt_formulation}

Our trajectory optimizer assumes it is given some goal pose for $O_{\text{new}}$, denoted $T^*_{\text{new}}$, along with a set of objects $\mathcal{G}$ in collision with the goal pose. To solve this trajectory optimization problem we use the iLQR algorithm \cite{li2004iterative,tassa2012synthesis}, along with MuJoCo \cite{todorov2012mujoco} to simulate contact interactions. The algorithmic details of iLQR are outside the scope of this paper, briefly, iLQR solves the following non-linear trajectory optimization problem:
\begin{equation}
\begin{aligned}
\min_{\mathbf{u}_{0:T-1}} \quad & \ell_f(\mathbf{x}_T) + \sum^{T-1}_{t=0} \ell(\mathbf{x}_t, \mathbf{u}_t) \\
\text{subject to} \quad & \mathbf{x}_{t+1} = f(\mathbf{x}_t, \mathbf{u}_t)\\
& \mathbf{x}_0 = \mathbf{x}(0)
\end{aligned}
\end{equation}
where, $\mathbf{x}_t \in \mathbb{R}^{n_x}$, $\mathbf{u}_t \in \mathbb{R}^{n_u}$ are the state and control vectors respectively at time-step $t$ along a trajectory. The state vector contains the positional and velocity components of the robot joints as well as all objects within the container ($n_x = n + 6\cdot|\mathcal{O}|$). The control vector is simply the actuators of the robot arm ($n_u=7$). $f(\mathbf{x}_t, \mathbf{u}_t)$ is the non-linear system dynamics. $T$ is the optimization horizon. $\ell(\mathbf{x}_t, \mathbf{u}_t)$ and $\ell_f(\mathbf{x}_T)$ are the state and terminal cost functions respectively.

iLQR requires cost derivatives about the nominal trajectory to compute improvements to the trajectory. Howell et al. \cite{howell2022predictive} introduce an efficient method to specify complicated cost terms and easily compute their derivatives. They implement their cost function in the following form:
\begin{equation}
    \ell(\mathbf{x}_t, \mathbf{u}_t) = \sum_{i=1}^{K} w_i \cdot n_i(\mathbf{r}_i(\mathbf{x}_t, \mathbf{u}_t)),
\end{equation}
where $K$ is the number of cost terms, $w_i$ is a scalar weight term, $n_i$ is some twice differentiable function and $\mathbf{r}_i(\mathbf{x}_t, \mathbf{u}_t)$ is a residual function that is `small when the task is solved'. This cost specification enables an easy method of specifying task costs where the cost derivatives can be approximated by computing the first order residual derivatives via finite-differencing and approximating the second order residuals via the Gauss-Newton approximation. The cost derivatives can then be computed via the chain-rule.

\subsubsection{Residuals}

We specify the following residuals to incite the desired behaviour of our robotic manipulator:

\begin{itemize}
    \item \textbf{Position X (POS X)}: Position difference between the current position of $O_{\text{new}}$ ($T_{\text{new}}$) and the desired pose ($T_{\text{new}}^*$) along the world $x$-axis: $t_x(T_{\text{new}}) - t_x(T^*_{\text{new}})$.
    \item \textbf{Position Y (POS Y)}:  
    Similar, but along the $y$-axis.
    \item \textbf{Position Z (POS Z)}: Similar, but along the $z$-axis.
\end{itemize}
These three residuals simply incentivise moving $O_{\text{new}}$ towards the current goal pose $T_{\text{new}}^*$. These residuals are specified separately instead of simply penalising the 3D Euclidean distance so that different axes can be penalised differently. 

\begin{itemize}
    \item \textbf{Upright (UP)}: Rotational deviation of the object’s $z$-axis from the world $z$-axis: $\arccos(R_{\text{new}}[z] \cdot z_{world})$.
\end{itemize}
The upright residual ensures that the z-axis of the object is aligned with the z-axis of the world frame.

\begin{itemize}
    \item \textbf{End-Effector Force (EE FORCE)}: Magnitude of the 3D force vector at the end-effector tip: $\mathbf{f}_{ee}$.
\end{itemize}
We add a force sensor inside the physics simulator at the end-effector tip and penalise forces. 

\begin{itemize}
    \item \textbf{Robot Joint Limit (RJL)}: Joint limit avoidance for joint $i$ away from its upper and lower bounds ($\bar{\mathbf{q}}_i^l$ and $\bar{\mathbf{q}}_i^u$): $\cosh\!\big(\mathbf{q}_i - \tfrac{\bar{\mathbf{q}}_i^u + \bar{\mathbf{q}}_i^l}{2}\big) - 1$.
    \item \textbf{Robot Joint Velocity (RJV)}: Joint velocity penalty for joint $i$: $\dot{\mathbf{q}}_i$.
\end{itemize}
 \textbf{RJL} adds penalties when the robot begins to approach its joint limits. We use a $cosh$ function so that the cost is small when the joint is near its midpoint and then gets large as it approaches its joint limits. \textbf{RJV} simply penalises robot joint velocities, which helps to ensure that $O_{\text{new}}$ is not dropped due to inertial forces.
\begin{itemize}
    \item \textbf{Obstacle Repulsion (OR)}: Repulsion for obstacle $j$ in 
    $\mathcal{G}$ away from packing pose:
    $ e^{-\sqrt{(t_x(T_j) - t_x(T^*_{\text{new}}))^2+((t_y(T_{i}) - t_y(T^*_{\text{new}}))^2}} $.
\end{itemize}
Finally, for any objects that are in the collision list $\mathcal{G}$, an obstacle repulsion cost is specified which incentivizes the robot to displace the object away from the goal pose $T_{\text{new}}^*$. To prevent the robot focusing on displacing objects for too long during the task, we used a simple decay mechanism which caused the obstacle repulsion residual weights to reduce to zero over time:
\begin{equation}
    w_{obs,t} = \text{max} \left( 0, \quad w_{obs,0} - \frac{t \cdot w_{obs,0}}{Z}\right),
\end{equation}
where $w_{obs,t}$ is the weight of the obstacle repulsion residual at time-step $t$, $w_{obs,0}$ is the weighting initially and $Z$ is some number of time-steps for the weight to decay over.


\subsection{MPC System Design} \label{sec:MPC_system_design}

Performing contact-based trajectory optimization in simulation and executing open-loop on real robotic hardware often fails due to inaccuracies between real world contact and simulation. As such, it is generally necessary to perform Model Predictive Control (MPC) when executing trajectories on real robotic hardware. We implement our trajectory optimization formulation under a short-horizon asynchronous MPC scheme to enable the real robotic system to account for these deviations.

Contact-based MPC on real robotic hardware generally requires two additional considerations: (i) System state estimation, so that successive optimization iterations can be performed from the current state of the system; and (ii) a low-level robotic controller that tracks the optimized trajectory, to account for the slower optimization times typically encountered during contact-based MPC. We chose a joint PD controller with feed forward torque as our low-level controller:

\begin{equation}
    \tau_a = \tau_{ff} + \mathbf{K}_p(\mathbf{q} - \mathbf{q}_d) + \mathbf{K}_d(\mathbf{\dot{q}} - \mathbf{\dot{q}}_d),
\end{equation}
where $\tau_a$ are the torques sent to the robot, $\tau_{ff}$ are the feed forward torques computed from MPC. $\mathbf{q}_d$ and $\mathbf{\dot{q}}_d$ are the desired joint positions and velocities from MPC, where the difference between these and the actual values are multiplied by some gain matrices $\mathbf{K}_p$, $\mathbf{K}_d$. An overview of the robotic control system can be seen in Fig. \ref{fig:MPC_system_diagram}.

\subsubsection{System state estimation}

Perception is challenging in such environments, due to the presence of clutter and also due to the robot, and the inserted object, obscuring the environment from the view of cameras. To alleviate this issue, we combined our object pose tracking system with a physics-based state estimator \cite{xu2025tracking}.

The core overview of our physics-based tracking system was that it would primarily use poses provided by some vision system for objects, however, when objects in the container could not reliably be observed (due to obstruction by the robot or inserted object) it would use physics estimates based on the robot motion and last known poses of objects to predict how objects within the container would move. In this work, we use an OptiTrack vision system to track the poses of objects within the container.



\subsubsection{Task Stages}

We divide the task into three stages: \textbf{PRE-PACK}, \textbf{INSERT}, and \textbf{FINE-TUNE}. Each stage employs different cost weights (as can be seen in Table \ref{table:residuals}), and in some cases different residual targets, to elicit distinct behaviours from the manipulator. We assume we have been given some desired packing pose $T_{\text{new}}^{\text{pack}}$. The stages are described as follows:

\noindent \textbf{PRE-PACK}: The manipulator first positions $O_{\text{new}}$ in a ``pre-packing pose'' $T^{\text{pre}}_{\text{new}}$, which shares the same $(x,y)$ location as the packing pose $T^{\text{pack}}_{\text{new}}$ but is offset by a positive clearance $\Delta z$ in the vertical direction:
\begin{equation}
    t(T^{\text{pre}}_{\text{new}}) = \big(t_x(T_{\text{new}}^{\text{pack}}),\, t_y(T_{\text{new}}^{\text{pack}}),\, t_z(T_{\text{new}}^{\text{pack}}) + \Delta z\big).
\end{equation}
During this stage, $T^*_{\text{new}}$ is set to the pre-packing pose $T^{\text{pre}}_{\text{new}}$. As no objects are in collision with $T_{\text{new}}^{\text{pre}}$, the collision list $\mathcal{G}$ is empty. This stage prepares the robot for placement by establishing a favourable configuration before interacting with the clutter. The stage is complete once the desired pre-packing pose is achieved within some distance threshold.

\noindent \textbf{INSERT}: This stage involves displacing objects within the container using non-prehensile manipulation so that $O_{\text{new}}$ can be lowered towards the desired height of the packing pose. $T_{\text{new}}^*$ is set as the goal packing pose $T_{\text{new}}^{\text{pack}}$. Generally, $T_{\text{new}}^{\text{pack}}$ will intersect with other objects in the container, these objects will be added to the collision list $\mathcal{G}$. The manipulator must clear sufficient space to accommodate $O_{\text{new}}$, but importantly, it does not need to reach the exact 2D position of $T^\text{pack}_\text{new}$, instead prioritising reaching the base of the container (determined by some distance threshold).

\noindent \textbf{FINE-TUNE}: By this stage, the object has reached the base of the container, however the object is not necessarily within some 2D distance threshold of $T_{\text{new}}^\text{pack}$. This final stage refines the placement by moving $O_{\text{new}}$ towards the target packing pose until it lies within a specified tolerance.

\begin{table*}[b] 
\caption{Weight modifiers for the specified residuals for each of the three task stages.}
\centering
\setlength{\tabcolsep}{2.7pt} 
\begin{tabular}{l c c c c c c c c} 
\hline
\textbf{Task Stage} & \textbf{POS X} & \textbf{POS Y} & \textbf{POS Z} & \textbf{UP} & \textbf{EE FORCE} & \textbf{RJL} & \textbf{RJV} & \textbf{OR} \\
\hline
\textbf{PRE-PACK} & \makecell{$w=2$ \\ $w_f=100$} & \makecell{$w=2$ \\ $w_f=100$} & \makecell{$w=0.1$ \\ $w_f=50$} & \makecell{$w=0.1$ \\ $w_f=20$} & \makecell{$w=(0, 0, 0)$ \\ $w_f=(0, 0, 0)$}  & \makecell{$w=0.1$ for all joints \\ $w_f=0.1$ for all joints} & \makecell{$w=0.02$ for all joints \\ $w_f=1$ for all joints} & \makecell{$w=0$ \\ $w_f=0$} \\
\hline
\textbf{INSERT} & \makecell{$w=1$ \\ $w_f=100$} & \makecell{$w=1$ \\ $w_f=100$} & \makecell{$w=1$ \\ $w_f=200$} & \makecell{$w=0$ \\ $w_f=0$} & \makecell{$w=(1, 1, 1)$ \\ $w_f=(1, 1, 1)$} & \makecell{$w=0.1$ for all joints \\ $w_f=0.1$ for all joints} & \makecell{$w=0.15$ for joints 1,3; 0.02 others \\ $w_f=0.15$ for joints 1,3; 0.02 others} & \makecell{$w=1$ \\ $w_f=1$} \\
\hline
\textbf{FINE-TUNE} & \makecell{$w=1$ \\ $w_f=200$} & \makecell{$w=1$ \\ $w_f=200$} & \makecell{$w=0.01$ \\ $w_f=20$} & \makecell{$w=0.1$ \\ $w_f=20$} & \makecell{$w=(0, 0, 0)$ \\ $w_f=(0, 0, 0)$} & \makecell{$w=0.1$ for all joints \\ $w_f=0.1$ for all joints} & \makecell{$w=0$ for all joints \\ $w_f=0$ for all joints} & \makecell{$w=0$ \\ $w_f=0$}\\
\hline
\end{tabular}
\label{table:residuals}
\end{table*}
\begin{figure}[t]
    \centering
    \scriptsize
    \def\svgwidth{0.98\columnwidth}{\input{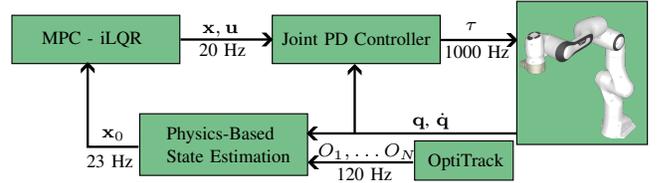}}
    \caption{Control system architecture for MPC on real robotic hardware.}
    \label{fig:MPC_system_diagram}
\end{figure}

\subsection{Placement Planner} \label{sec:placement_planner}

\begin{algorithm}[t!]
\caption{Placement Planner}
\label{alg:placement_planner}
\begin{algorithmic}[1]
\Require Initial system state \( \mathcal{P}_0 \), number of samples \( S \), container pose $T_{container}$, width $W$ and length $L$. Object to be packed $O_{\text{new}}$
\Ensure Selected pose $T^{\text{pack}}_{\text{new}}$ for target object $O_{\text{new}}$

\State Initialise cost list  $\mathcal{C} \gets [~]$ , pose list $ \mathcal{P} \gets [~] $, object collision list $\mathcal{G} \gets [~]$

\For{$i = 1$ to $S$}
    \State $ \mathcal{P}_i \gets \mathcal{P}_0 $
    \State $T^i_{\text{new}} \gets \texttt{SamplePose($T_{\text{container}}, W, L$)}$
    \State $ \mathcal{P}_i \gets \texttt{Teleport}(O_{\text{new}}, T_{\text{new}}^i, \mathcal{P}_i) $
    \For{$t = 1$ to $M$}
        \State $ \mathcal{P}_i \gets \phi(\mathcal{P}_i) $ \Comment{Integrate system physics}
    \EndFor
    \State $T_{\text{new}}^{i,\text{settled}} \gets \texttt{GetPose}(O_{\text{new}}, \mathcal{P}_i)$ 
    \State $c_i \gets J(\mathcal{P}_0, \mathcal{P}_i)$

    \State $ \mathcal{P}_i \gets \texttt{Teleport}(O_{\text{new}}, T_{\text{new}}^{i}, \mathcal{P}_i) $
    \State $ \mathcal{G} \gets \texttt{GetCollisions}(\mathcal{P}_i)$
    
    \State $\mathcal{C} \gets c_i$
    \State $\mathcal{H} \gets T_{\text{new}}^{i,\text{settled}}$
\EndFor

\State $ k \gets \arg\min_i \mathcal{C}[i] $
\State $ T^{\text{pack}}_{\text{new}} \gets \mathcal{H}[k] $
\State \Return $T^{\text{pack}}_{\text{new}}$ , $\mathcal{G}[k]$ 
\end{algorithmic}
\end{algorithm}

\begin{figure}[t]
  \centering
  \includegraphics[width=0.98\columnwidth]{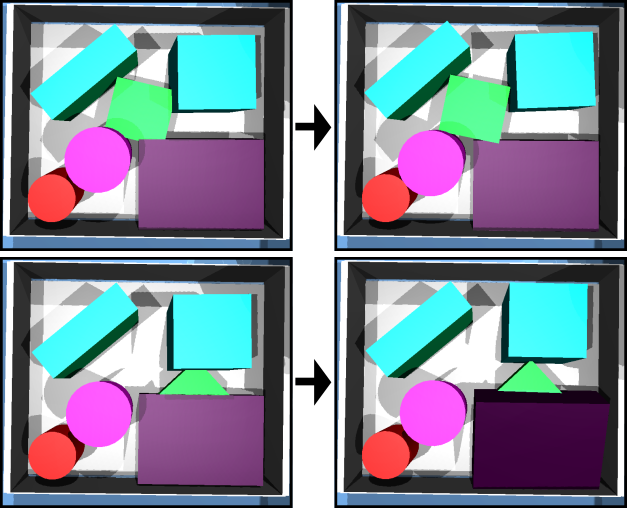}
  \caption{Two visualisations of the same packing scene with different sampled packing poses. The green object is the one to be packed, all other objects are present in the container before packing. The left image is the first frame when the packed object is teleported inside the container and the right image is after the physics simulator has been integrated $M$ times. The top row shows the best packing pose found for this scene, as it has minimal disturbance to several objects surrounding it. The bottom row incurs a much larger cost from Eq. \ref{eq:placement_cost} as it is unable to displace the purple object inside the container plane and instead, causes the object to move upwards outside of the container, implying this would be an unsuitable target packing pose.}
  \label{fig:placement_packing_snapshots}
\end{figure}

As discussed in Sec. \ref{sec:traj_opt_formulation} our trajectory optimization formulation requires a packing pose $T_{\text{new}}^\text{pack}$, as well as a list of any objects in collision with the packing pose $\mathcal{G}$. To this end, we propose a high-level placement planner to identify a suitable pose inside the container to place $O_{\text{new}}$. The objective of the placement planner is to find a pose which minimises disturbance to the scene.

The placement planner samples possible placement poses $T^{i}_{\text{new}}$ for $O_{\text{new}}$ within the bounds of the container along its bottom plane. To identify the packing viability of this placement pose, $O_{\text{new}}$ is teleported to that pose in the physics simulator (most likely resulting in $O_{\text{new}}$ being in collision with other objects in the container). The physics simulator is then integrated $M$ times to resolve the collisions inside the container. The use of MuJoCo's soft contact model ensures that the simulation remains stable, despite penetrations. Let $\mathcal{P} = \{T_1, T_2, \ldots,T_N \}$ represent the configuration of all objects inside the container. $\mathcal{P}_0$ is the initial configuration of the container, $\mathcal{P}_i$ is the state of the system after teleporting $O_{\text{new}}$ to the sampled pose $T^{i}_{\text{new}}$ and $\mathcal{P}_i^{\text{settled}}$ is the state of the scene after the physics simulator has been integrated $M$ times. We define $c_i$ as the evaluated cost of sampled pose $i$ and $\mathcal{C} = \{ c_1, c_2, \ldots c_S\}$ is the list of sampled costs. Let $\mathcal{H} = \{ T^{\text{pack},1}_{\text{new}}, T^{\text{pack},2}_{\text{new}}\ldots T^{\text{pack},S}_{\text{new}} \}$ represent the list of sampled packing poses for $O_{\text{new}}$.

To evaluate the packing viability of a sampled packing pose $T^{i}_{\text{new}}$, we define a cost function $J(\mathcal{P}_0, \mathcal{P}_i^{\text{settled}})$ which evaluates how much disturbance was experienced inside the container due to the packing pose $T^{i}_{\text{new}}$:
\begin{align}
\label{eq:placement_cost}
J(\mathcal{P}_0,\mathcal{P}_i^{\text{settled}})
&=
\sum_{j=1}^{N}
\Big\|
  W^{\top}
  \big(t(T^{i,\text{settled}}_j) - t(T^{i}_j)\big)
\Big\|_2 \\
&+ \beta\mathcal{K} (\mathcal{P}_i^{\text{settled}}), \nonumber
\end{align}
where $W = (w_x,\;w_y,\;w_z)$ are weights which penalise deviations in the x, y and z axes. The function $\mathcal{K}(\mathcal{P}_i)$ penalises any lasting penetrations of objects inside the container after $M$ physics integrations and $\beta$ was a weight modifier. 

After $S$ samples have been taken and evaluated, we choose the sample with the lowest cost as per Eq. \ref{eq:placement_cost}. The placement planner returns both an ideal goal packing pose $T_{\text{new}}^*$, as well as a list of objects inside the container which are in collision with $O_{\text{new}}$ at the goal packing pose $\mathcal{G}$. An overview of the proposed placement planner is shown in Alg. \ref{alg:placement_planner}.

We visualise an example packing scene in simulation with two sampled poses in Fig. \ref{fig:placement_packing_snapshots}.  The top row of the figure shows the best packing pose found for this configuration of objects inside the container, whereas the bottom row shows a bad packing pose as it was not able to simply displace any objects within the 2D plane of the container and instead the object was displaced upwards out of the container. As was discussed earlier, vertical displacements are punished heavily as they suggest that such a packing pose is infeasible by simply perturbing the object inside the container's 2D plane.

\section{EXPERIMENTS} \label{sec:experiments}

In this section, we present results demonstrating the performance of our methods on both real robotic hardware and in simulation. As is commonly observed when transferring systems from simulation to hardware, success rates decrease in the real-world setting. We analyse this drop and highlight several contributing factors, which suggest directions for future work. To better understand the role of the different components of our method, we conduct ablation experiments by comparing our full method against two baselines. The first baseline omits the placement planner, while the second ignores the physics of other objects inside the container during trajectory optimization. 

Sec. \ref{sec:methods_and_baselines} formally defines our method and baselines. In Sec. \ref{sec:implementation_details}, we provide relevant implementation details. Sec. \ref{sec:real_world_experiments} presents the overall results from real robot experiments and we refer the reader to the accompanying video for demonstrations of real robot executions. We perform follow up simulation experiments in Sec. \ref{sec:simulation_experiments} to compare the performance between simulation and the real world. Finally, Sec.~\ref{sec:hardware_failure_modes} analyses common failure modes in real-robot execution.

\subsection{Methods and Baselines} \label{sec:methods_and_baselines}

We evaluate our full method as proposed in Sec. \ref{sec:method} which we name \textbf{PackItIn}. We also evaluate two baselines where we remove one of the two aspects of our proposed method:

\begin{itemize}
    \item \textbf{Uninformed Pose}: This baseline does not use the placement planner as described in Alg. \ref{alg:placement_planner}, instead it simply randomly chooses a packing pose inside the container. Importantly it still adds any corresponding objects in collision with the packing pose to the collision list $\mathcal{G}$.
    \item \textbf{Uninformed Physics}: This baseline still receives an informative packing pose from Alg. \ref{alg:placement_planner}, however, during trajectory optimization, it does not consider the physical interactions between objects in the container and $O_{\text{new}}$. This is achieved by removing the objects from the simulated container and not tracking their poses from the vision system.
\end{itemize} 

\subsection{Implementation details} \label{sec:implementation_details}

The optimization horizon $T$ used for MPC was 0.6 seconds and the model time-step $\Delta t$ was 0.006 seconds, resulting in an optimization horizon of 100 time-steps. To speed up trajectory optimization iterations and make this problem computationally tractable, we employed a naive set interval dynamics derivative skipping approach \cite{russell2023adaptive} to speed up the computational bottleneck of iLQR that comes from finite-differencing. 

As our system is model-based, we assume access to semi-accurate models of the real-world objects used in simulation. For our experiments, we selected five random objects commonly found in a local supermarket to serve as distractors inside the container. Each object was weighed and measured to obtain the necessary model parameters, and the set consisted of simple cylinders and boxes. We emphasise that this restriction to basic geometries was chosen purely for ease of implementation; in principle, our method is directly applicable to more complex object geometries. We model $O_{\text{new}}$ as rigidly connected to the pump tip.

Our experiments were performed on a 16 virtual core CPU
(11h Gen Intel(R) Core(TM) i9-11900@2.50GHz) with 128
GB RAM. A Franka-Emika Panda robotic manipulator was used to perform the packing operations using a vacuum pump attachment, namely the Schmalz Cobot Pump. As was shown in Fig. \ref{fig:MPC_system_diagram}, we employ a low-level PD with feed forward torque controller to assist with joint-level tracking. We use positional gain values of $(100, 100, 100, 100, 100, 60, 60)$ and velocity gain values of $(75, 75, 75, 75, 20, 12, 12)$.

With regards to the placement planner, we set the number of sampled poses in the container for the placement planner $S$ to be 300 as we found that sufficient to find an informative packing pose. When evaluating sampled packing poses we used the following weight values for Eq. \ref{eq:placement_cost}: $w_x = 1, w_y = 1, w_z = 10, \beta = 1000$. The reasons behind these choices were to penalise packing poses that were either not resolved in $M$ ($M=30$) physics integrations (implying that the packing pose is difficult to achieve) as well as penalise packing poses that could only be achieved by displacing objects upwards away from the container's base plane (as this would be dynamically difficult to achieve). 

\subsection{Hardware Experiments} \label{sec:real_world_experiments}

\begin{table*}[t] 
\centering
\caption{Packing on real hardware results averaged over 40 trials. Numeric results are mean value $\pm$ standard deviation.}
\begin{tabular}{l c c c c c c}
\hline
\makecell{\textbf{Method}} & \textbf{Success} & \textbf{Force Failure} & \textbf{Object Detached} & \textbf{Exec Time (s)} & \textbf{Dist Error (cm)} & \textbf{EE Force (N)} \\
\hline
\makecell{PackItIn} & $28 / 40$ & $3 / 40$ & $5/40$ & $10.74 \pm 6.14$ & $5.0 \pm 3.7$ & $5.36 \pm 1.40$\\
\makecell{Uninformed Physics} & $8 / 40$ & $2 / 40$ & $12/40$ & $16.44 \pm 5.68$ & $6.7 \pm 5.7$ & $7.52 \pm 2.31$ \\
\makecell{Uninformed Pose} & $7 / 40$ & $4 / 40$ & $8/40$ & $17.85 \pm 4.18$ & $7.7 \pm 4.1$ & $4.97 \pm 1.95$ \\
\hline
\end{tabular}
\label{table:real_robot_results}
\end{table*}

\begin{figure}
    \centering
    \includegraphics[width=0.98\columnwidth]{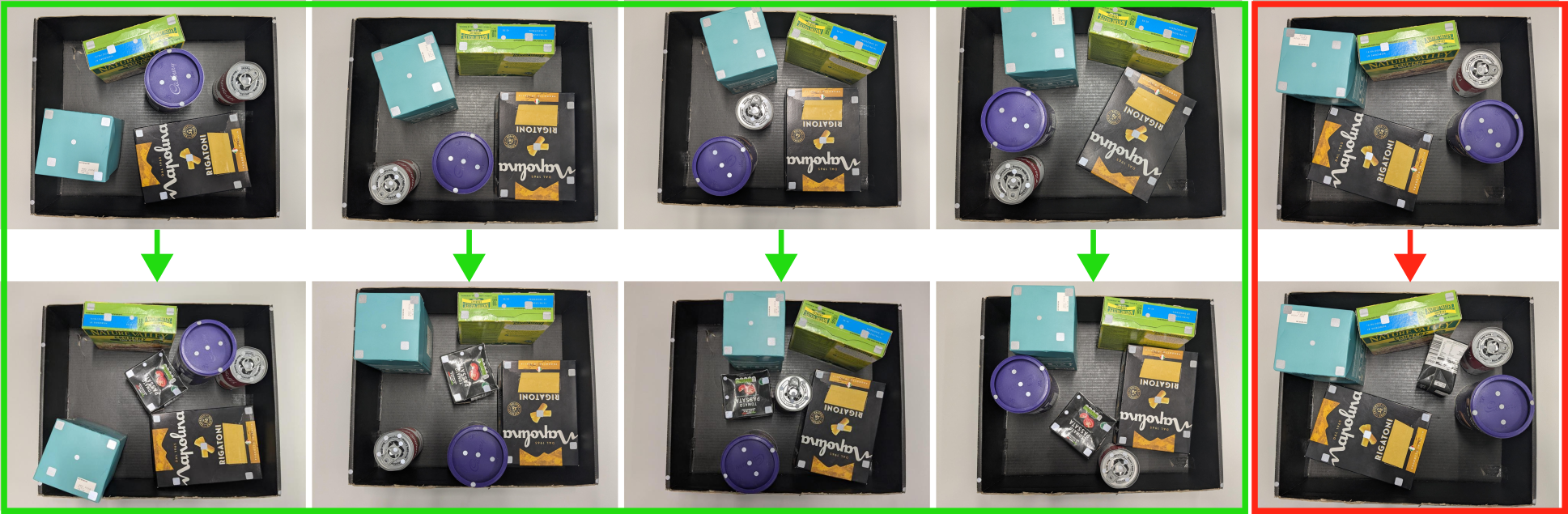}
    \caption{Overhead views of 5  of the 40 packing scenes used during hardware experiments. Top row shows the state of the scene before packing and the bottom row shows the state of the scene after the object (tomato passata) was packed inside the container. Four of these trials were successful (green outline) and one was unsuccessful (red outline). The unsuccessful case was an object detached failure, hence the object is in a different orientation.}
    \label{fig:testing_scenes}
\end{figure}

We first present experimental results of our method and the two baseline variants on real robotic hardware. A total of 40 evaluation scenes were randomly generated, each created by shaking the container and inserting objects so as to intentionally reduce the available free space, thereby increasing the difficulty of inserting $O_{\text{new}}$. For a visual reference of these experimental setups and executions, please see the accompanying video. To ensure fairness in evaluation, we photographed each randomly generated scene and then carefully reconstructed it so that all three methods were tested under the same configuration of objects within the container. Fig. \ref{fig:testing_scenes} shows 5 example scenes used for testing, the top row of this figure shows the state of the scene before robot execution and the bottom row shows the state of the scene after the object was packed.

Each trial was executed until one of four exit conditions was met: (1) task completion, when $O_{\text{new}}$ was successfully placed at the target packing pose, within a distance threshold of 1.5 cm; (2) task timeout, after twenty seconds of robot execution without meeting any other condition; (3) object detachment, when the object prematurely separated from the pump tip during execution; or (4) force failure, when external forces triggered a safety condition that halted robot operation. Notably, we deem task success as the system reaching the third task stage (FINE-TUNE) as the object has been successfully inserted into the container at this point. The system continues running after this until the object is located within some small distance threshold of the desired packing pose, or until one of the failure modes above occur.

The overall results of these experiments are presented in Table \ref{table:real_robot_results}. The task success, force failure and object detached exit modes are denoted in the table (with the remainder of trials exiting due to a task timeout). In addition to these values, we show the average task execution time (not counting trials that exited early due to force failure or object detached failure); 3D distance error between the final pose of $O_{\text{new}}$ and the goal packing pose; as well as the average external force readings experienced at the end-effector.

PackItIn achieves a 70\% success rate over the real robotic experiments, in comparison to the other two baselines which only achieved a 17.5\% and 20\% success rate. This substantial difference in success rates shows the importance of both aspects of our proposed method, as both baselines perform equally poorly in comparison.

Force failure and object detached errors occur in a total of 20\% of the failures for our PackItIn method. Force failures occurred when trying to displace an object away from the desired packing pose, when our trajectory optimization formulation instructed the robot to move in such a way that incurred large external forces due to object jamming. Notably, our method to penalise simulated EE forces appeared to reduce this effect during early testing, but was not able to completely eliminate this failure mode. The primary reason for object detached failure modes was when our assumption that the object was rigidly attached to the pump tip was violated significantly. Due to the minimalist pump we used, the object could rotate about the pump tip when it was used displace objects within the scene, if this rotation became too significant it could result in the object becoming removed from the pump tip.

There are several possible reasons for the various failure modes on real robotic hardware. These could be related to issues with our perception system, model mismatch between the physics simulator and the real physics, or they could be caused by issues in the motion planning itself, even without taking into account sim2real issues. We perform simulation evaluations to first establish the efficacy of our system in an ideal simulated environment.

\subsection{Simulation Experiments} \label{sec:simulation_experiments}

In simulation, we generated 100 random scenes programmatically using the same five objects as in the real-robot experiments. The evaluation metrics are identical to those in Section~\ref{sec:real_world_experiments}, and the overall results are summarised in Table~\ref{table:simulation_results}. The proposed \textbf{PackItIn} method achieved the highest success rate (93\%), outperforming both baselines, while also yielding the lowest average execution time and final 3D distance error.

Two observations stand out from these results. First, success rates are substantially higher in simulation than on hardware, even for the naïve baselines: Uninformed Physics reached 78\% and Uninformed Pose 60\%. These inflated numbers stem from the compliant nature of the simulation environment. For example, when provided with an unfavourable packing pose, the simulated robot often found a way to reach the base of the container by exploiting inaccuracies inside the physics simulator. Similarly, the Uninformed Physics baseline succeeded in many trials simply by moving directly towards the target pose. This behaviour is facilitated by the soft-contact model in the MuJoCo simulator, which permits interpenetration between bodies and gradually resolves collisions.

Second, in simulation all failures were exclusively due to task timeouts. Unlike on hardware, no force-based failure condition was required, and detachment never occurred because the simulated robot assumed the object was rigidly fixed to the pump tip. \textbf{PackItIn} failed in 7\% of simulated trials, all due to timeouts, which is comparable to the 10\% timeout rate observed on real hardware. These findings suggest that the primary factors reducing success rates on hardware are the additional failure conditions—namely force-based exits and object detachment—that are absent in simulation.

\begin{table}[t]
\centering
\caption{Simulation packing through clutter results.}
\setlength{\tabcolsep}{5pt} 
\begin{tabular}{l c c c c}
\hline
\makecell{\textbf{Method}} & \textbf{Success} & \textbf{Exec Time (s)} & \textbf{Dist (cm)} & \textbf{Force(N)} \\
\hline
\makecell{PackItIn} & $93 / 100$ & $8.1 \pm 5.0$ & $2.0 \pm 1.3$ & $0.35 \pm 0.08$\\
\makecell{Uninf. Physics} & $78 / 100$ & $10.2 \pm 6.4$ & $3.8 \pm 3.9$ & $0.24 \pm 0.09$ \\
\makecell{Uninf. Pose} & $60 / 100$ & $15.7  \pm 5.6$ & $4.2 \pm 3.0$ & $0.34 \pm 0.20$\\
\hline
\end{tabular}
\label{table:simulation_results}
\end{table}

\subsection{Hardware Failure Cases} \label{sec:hardware_failure_modes}



Finally, returning to the real hardware experiments, we identified three primary factors that contributed to task failures. These were quantified using metrics accumulated over each trial and normalised by the total trial duration. The first metric captured the deviation between simulated and actual object positions during execution. The second metric quantified model mismatch in the orientation of the inserted object $O_{\text{new}}$, specifically measuring the unmodeled rotation of the object about the pump tip. The final metric measured failures of the perception system, i.e., the frequency with which object poses had to be estimated via physics-based predictions rather than direct detection by OptiTrack. 

\begin{figure}[t]
    \centering
    \scriptsize
    \def\svgwidth{0.98\columnwidth}
    \scalebox{1}{\input{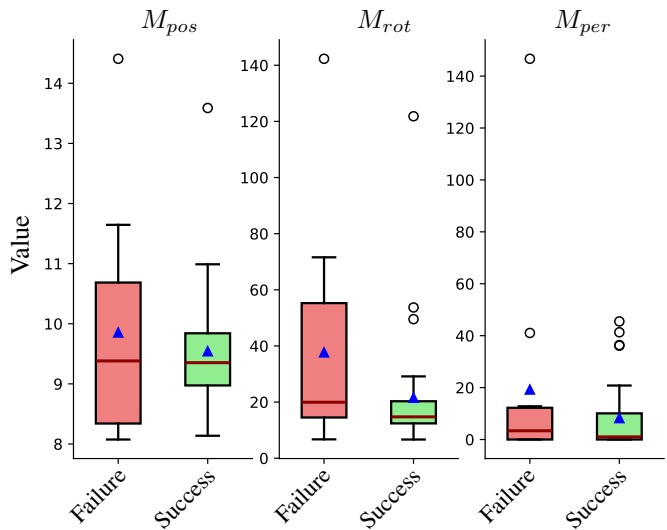}}
    \caption{Box plots of the metrics defined in Eqs.~\ref{eq:pos_error}--\ref{eq:per_error} for PackItIn trials. Each box indicates the interquartile range (IQR), with whiskers extending to 1.5$\times$IQR. Maroon line is median and blue triangle is mean value.}
    \label{fig:failure_modes_boxplots}
\end{figure}

Formally, these metrics were defined as follows:
\begin{equation} \label{eq:pos_error}
M_{\text{pos}} = \frac{1}{T_EN} \sum_{t=1}^{T_E} \sum_{i=1}^{N} 
\left\| t(T)^a_t - t(T)^s_t \right\|_2 ,
\end{equation}
\begin{equation} \label{eq:rot_error}
M_{\text{rot}} = \frac{1}{T_E} \sum_{t=1}^{T_E} 
\arccos\!\left( R_{\text{new},t}^a \cdot R_{\text{new},t}^s  \right) ,
\end{equation}
\begin{equation} \label{eq:per_error}
M_{\text{per}} = \frac{1}{T_E} \sum_{t=1}^{T_E} \sum_{i=1}^{N} 
\mathbf{1}_{\{\text{physics estimate used for object } i \text{ at } t\}} ,
\end{equation}
where $T_E$ is the execution time of the trial, $N$ is the number of objects within the container, $t(T)^a_t$, $t(T)^s_t$ are the actual versus simulated positions of object $i$ at timestep $t$. $R_{\text{new},t}^a$, $R_{\text{new},t}^s$ were the actual versus simulated rotations of $O_{\text{new}}$ at timestep $t$. Finally $\mathbf{1}_{\cdot}$ represents a  binary indicator, returning 1 if the condition is true (OptiTrack perception failure).

Fig.~\ref{fig:failure_modes_boxplots} presents the results of these metrics across the PackItIn trials, comparing successful and unsuccessful runs (force failure, object detachment, or task timeout). In all cases, the unsuccessful trials exhibited higher average values for all three metrics, with the largest discrepancy observed in $M_{rot}$. This suggests that rotational misalignment of $O_{\text{new}}$ was the primary driver of task failure in our system.

\section{CONCLUSION}

We have proposed a packing system capable of inserting new objects into partially filled containers by leveraging non-prehensile manipulation to create space during the packing process. Unlike conventional approaches that separate rearrangement and packing into distinct steps, our system performs both in a single action. The entire pipeline is physics-aware: we formulate packing as a trajectory optimisation problem and employ a general-purpose physics simulator to simulate contact interactions; we introduce a high-level placement planner that simulates physical interactions within the container to identify feasible packing poses; and we incorporate a physics-aware perception module that uses physics predictions for object pose estimation when objects inside the container are occluded from our vision system.

We validated our system on real robotic hardware and achieved a 70\% success rate for our overall method. We analyse the failure cases of our system and have determined that three main aspects need investigating to improve the overall success rate of the system: more accurate perception, improved modelling about the pump tip and robust physics estimates for motion planning.

To bring this system into industrial settings, several challenges need to be overcome. Firstly, a more generalisable vision system need to replace the marker based perception system in this work. Secondly, a method is needed to instantiate novel objects within the model based on vision of the scene (and possibly robot interaction). Finally, scaling to higher-levels of clutter will require dimensionality reduction techniques \cite{russell2024online}.

\bibliographystyle{IEEEtran}
\bibliography{References.bib}


\end{document}